\documentclass{article}
\usepackage{booktabs}
\usepackage{caption}
\usepackage{color}
\usepackage{graphicx}
\usepackage{inconsolata}
\usepackage{makecell}
\usepackage{microtype}
\usepackage{multirow}
\usepackage[round]{natbib}
\usepackage{soul}
\usepackage{subcaption}

\usepackage[T1]{fontenc}
\usepackage[a4paper, top=2.5cm, bottom=2.5cm, left=2.8cm, right=2.8cm]{geometry}
\usepackage[colorlinks=true,citecolor=blue,linkcolor=blue,urlcolor=blue]{hyperref}
\usepackage[utf8]{inputenc}
\usepackage[normalem]{ulem}

\definecolor{green}{rgb}{0.52, 0.73, 0.4}
\definecolor{grey}{rgb}{0.83, 0.83, 0.83}
\definecolor{red}{rgb}{1.0, 0.6, 0.6}

\title{\textbf{Towards Contamination Resistant Benchmarks}}
\author{\textbf{Rahmatullah Musawi\thanks{This paper is adapted from the Master's thesis of Rahmatullah Musawi, under the supervision of LU Sheng.}, LU Sheng\thanks{Email: boblusheung@gmail.com}}}
\date{}

\begin{document}

\maketitle

\begin{abstract}

The rapid development of large language models (LLMs) has transformed the landscape of natural language processing. Evaluating LLMs properly is crucial for understanding their potential and addressing concerns such as safety. However, LLM evaluation is confronted by various factors, among which contamination stands out as a key issue that undermines the reliability of evaluations. In this work, we introduce the concept of \textit{contamination resistance} to address this challenge. We propose a benchmark based on Caesar ciphers (e.g., ``ab'' $\rightarrow$ ``bc'' when the shift is 1), which, despite its simplicity, is an excellent example of a contamination resistant benchmark. We test this benchmark on widely used LLMs under various settings, and we find that these models struggle with this benchmark when contamination is controlled. Our findings reveal issues in current LLMs and raise important questions regarding their true capabilities. Our work contributes to the development of contamination resistant benchmarks, enabling more rigorous LLM evaluation and offering insights into the true capabilities and limitations of LLMs.\footnote{The code and data are available at \url{https://anonymous.4open.science/r/contamination-resistant-E0D4}.}

\end{abstract}

\section{Introduction}

The advent of large language models (LLMs) has largely changed the field of natural language processing and many facets of daily life \citep{brown2020fewshot,chung2022scaling,chowdhery2023palm,dubey2024llama3}. Extensive research has asserted that these models possess ``human-like abilities'' such as reasoning \citep{kojima2022zeroshot,wei2022emergent,wei2022cot,bubeck2023sparks,hagendorff2023human,itzhak2024bias,xie2024adaptive}. However, LLM evaluation is often confronted by various factors, and contamination, also known as test set contamination or data leakage, stands out as a key issue. Given that LLMs are trained on vast web corpora, there is concern that their apparent reasoning skills are superficial, and that they are merely retrieving memorized information from their training data   \citep{sainz2023evaluation,dong2024generalization,jiang2024investigating,li2024task,mueller2024context,ravaut2024howmuch}.

A key strategy to address contamination is to create a dynamic benchmark that receives continuous updates, with new test instances added after a model's training cutoff \citep{jain2024livecodebench,roberts2024contamination,shabtay2024livexiv,white2024livebench,zhang2024gsm1k,mahdavi2025leveraging}. The effectiveness of this method rests on a critical assumption: the models in question have not been trained on these newly added instances. However, given the black box nature of current LLMs, especially commercial ones, it is almost impossible to determine whether developers have continued training their models after the cutoff using the updated instances, and, if so, when this occurred. As a result, the reliability and validity of the newly added data still remain questionable. Furthermore, creating new instances can be laborious, and maintaining a consistent level of difficulty between old and new instances is challenging, which is crucial for preserving the fairness of model performance comparisons across old and new instances.

\begin{figure*}
\centering
\begin{subfigure}{0.3\textwidth}
    \centering
    \includegraphics[width=3.4cm]{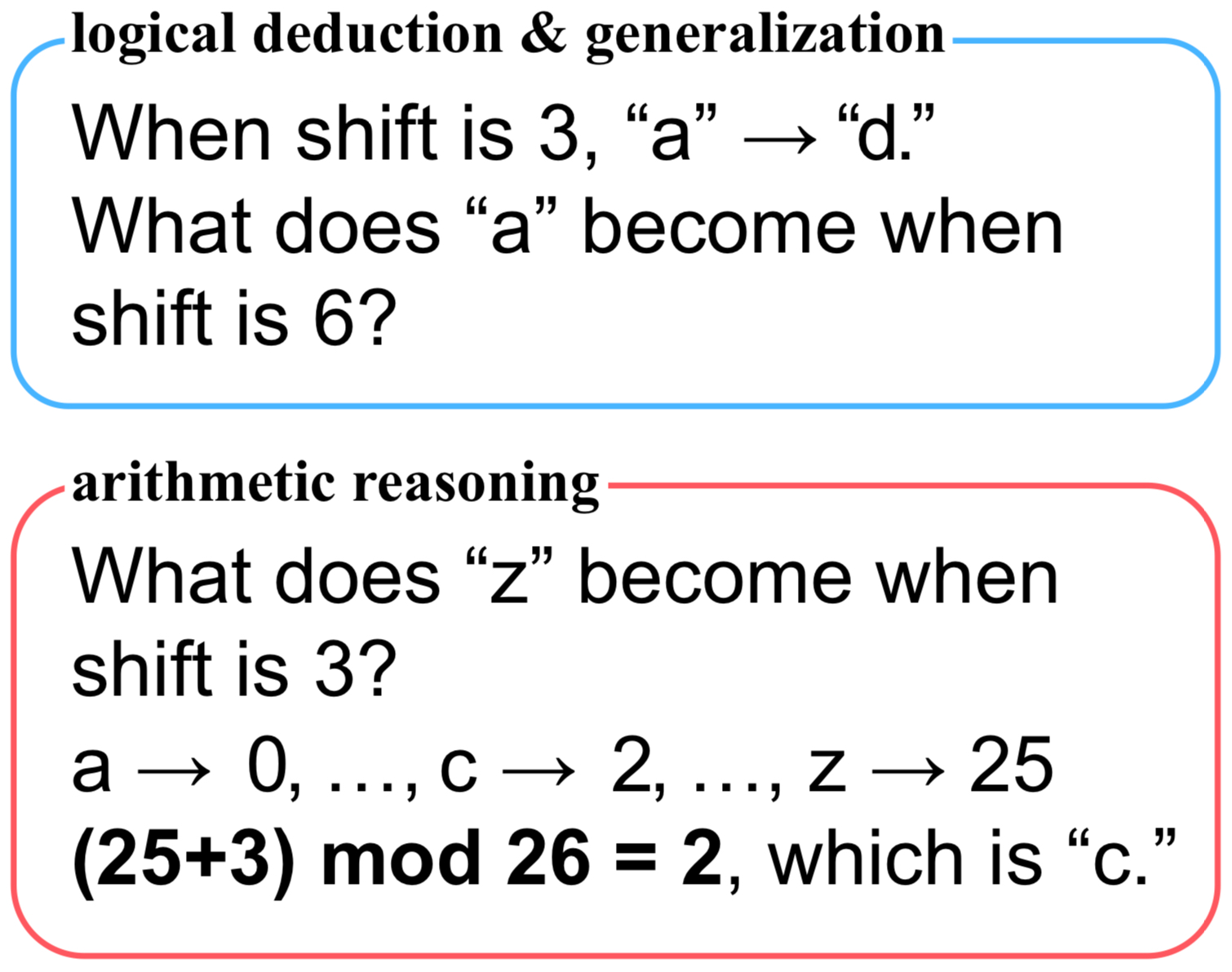}
    \caption{capabilities}
    \label{diagram_1}
\end{subfigure}
\begin{subfigure}{0.3\textwidth}
    \centering
    \includegraphics[width=4.8cm]{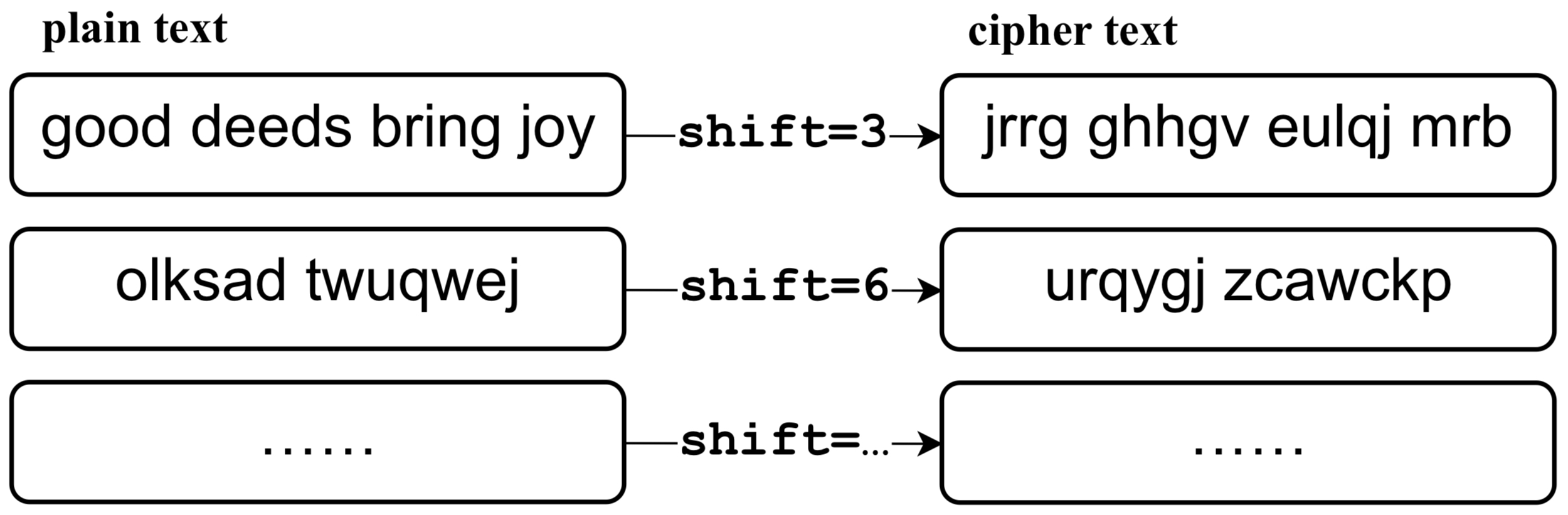}
    \caption{dynamism}
    \label{diagram_2}
\end{subfigure}
\hspace{1em}
\begin{subfigure}{0.3\textwidth}
    \centering
    \includegraphics[width=4.8cm]{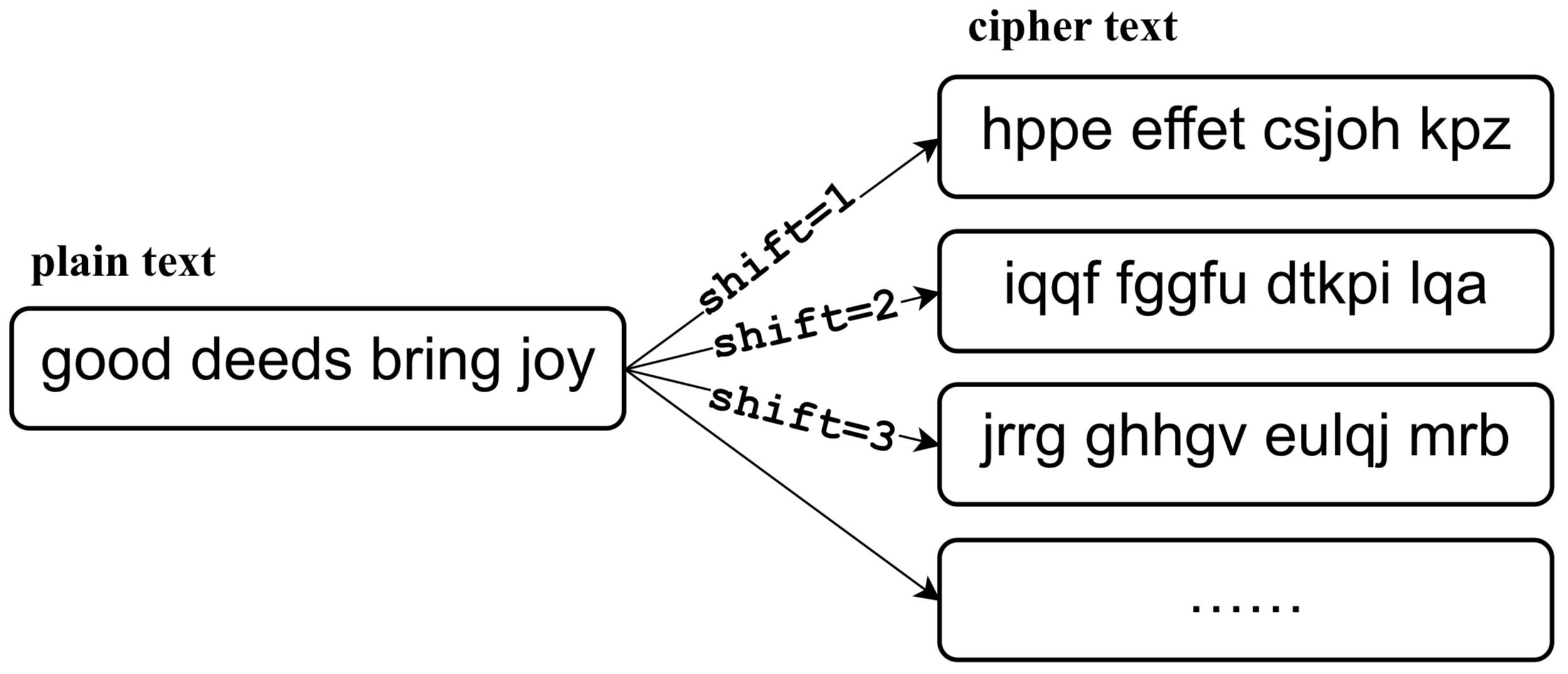}
    \caption{update-ability}
    \label{diagram_3}
\end{subfigure}
\caption{We propose the concept of \textit{contamination resistance} and establish a contamination resistant benchmark based on Caesar ciphers, which meets the following criteria: (a) this benchmark tests several capabilities, including logical deduction, arithmetic reasoning, and generalization; (b) this benchmark is dynamic, evolving over time so that an LLM can not memorize all possible queries, and the level of task complexity is consistent; (c) it is effortless to curate new test instances, making it more update-able than static benchmarks.}
\label{diagram}
\end{figure*}

Based on these considerations, we propose the concept of \textit{contamination resistance}. We argue that a contamination resistant benchmark should meet the following criteria: (a) it should assess certain capabilities of LLMs; (b) it should be dynamic, evolving over time to prevent contamination, and ensure that contamination provides no advantage in model performance; (c) while being dynamic, it should consistently assess the same set of capabilities and maintain the same level of task complexity to ensure fairness in evaluation over time; (d) the effort required to curate new instances should be minimal, making it more update-able than existing static benchmarks.

We show that a benchmark based on Caesar ciphers is an excellent example of a contamination resistant benchmark (see Figure \ref{diagram}). The Caesar cipher is a simple substitution cipher where each letter in a text is shifted some places down or up the alphabet. Encoding or decoding a Caesar cipher demands multiple capabilities, including logical deduction, arithmetic reasoning, and generalization. This benchmark is dynamic: users can generate an infinite number of unique texts, and it is unlikely that an LLM can memorize all possible queries. Despite its dynamic nature, this benchmark consistently evaluates the same underlying abilities and maintains the same level of task complexity. Moreover, this benchmark is extremely light weight. It requires little effort to generate new test instances and to perform inference.

Despite that solving a Caesar cipher is essentially a simple linear mapping task, and that the state-of-the-art LLMs perform exceptionally well on benchmarks that require much more complex reasoning abilities, they struggle with the Caesar cipher. We summarize our findings as \textbf{inconsistent competence of LLMs}, which add to the growing evidence that when contamination is controlled, LLMs are not performing as well as expected, highlighting the issue of contamination. Our findings also align with existing evidence of hallucination and the lack of generalization in LLMs.

Our contributions are summarized as follows:

\begin{itemize}
    \item We propose the concept of \textit{contamination resistance} and define the criteria a contamination resistant benchmark should meet.
    \item To establish a contamination resistant benchmark, we propose a benchmark based on Caesar ciphers and show that LLMs struggle with them when contamination is controlled.
    \item Our work paves the way for the development of contamination resistant benchmarks, enables more rigorous evaluation of LLMs, and sheds light on their true capabilities.
\end{itemize}

\section{Related work}

\subsection{Model capabilities}

The exceptional performance of LLMs on various tasks has led to claims that these models have gained ``human-like abilities'', such as reasoning \citep{kojima2022zeroshot,wei2022emergent,wei2022cot,bubeck2023sparks,saparov2023greedy,shi2023multilingual,webb2023analogical}. However, evidence that contradicts these claims exists, particularly concerning the phenomenon of hallucination \citep{bang2023hallucination,mckenna2023sources,mundler2023hallucination,zhang2023hallucination}. \citet{schaeffer2023mirage} argue that LLMs' abilities appear due to the choice of metric rather than fundamental changes in model behavior with scale. Another line of research suggests that LLMs rely more on memory than genuine reasoning abilities to solve certain tasks \citep{reynolds2021beyond,merullo2023word2vec,wang2023implicit,zheng2023chatgpt,li2024multilingual,lu2024sensitivity,mueller2024context}. \citet{tefnik2023context} introduce a conceptual learning method aimed at disentangling models' in-context learning ability from memorization. Their findings indicate that models rely heavily on their pre-trained knowledge than benefiting from the in-context concepts.

Whether LLMs truly possess ``human-like abilities'' has now gone beyond a technical problem. It affects how users interact with these models and has significant safety implications \citep{lu2024emergent,bengio2024managing}. By proposing the concept of contamination resistance, this work sheds light on the true capabilities of LLMs and helps promote better interaction between humans and LLMs.

\subsection{Model evaluation}

Evaluating LLMs properly is crucial to understand their potential as well as addressing concerns such as safety. The proliferation of benchmarks has led to a multifaceted evaluation process that covers a variety of abilities \citep{rajpurkar2018squad,wang2019superglue,cobbe2021gsm8k,srivastava2023beyond}. However, LLM evaluation is confronted by various factors, among which contamination stands out as one of the key issues. Contamination has become increasingly prominent given that current LLMs are trained on massive web corpora and are scaled up to billions of parameters. Furthermore, investigating potential data leakage is challenging because the training data are often closed source. These factors have severely undermined the reliability of evaluations \citep{sainz2023evaluation,balloccu2024leak,dong2024generalization,jiang2024investigating,ravaut2024howmuch,white2024livebench}. A growing body of research has revealed that LLMs perform better on problems that released before their training cutoff, and that some LLMs are overfitted to the testing sets of popular reasoning benchmarks \citep{eisenschlos2023winodict,li2024task,roberts2024contamination,zhang2024gsm1k}.

A straight forward approach to address contamination is to ensure that the LLM has never been exposed to the evaluation data during training. \citet{chandran2024private} introduce the notion of private benchmarking, where the evaluation data remain confidential and only the evaluation results are revealed. However, this approach may raise concerns regarding transparency and reproducibility. Another line of work focuses on dynamic benchmarking which features frequently updated questions from recent information sources introduced after a model's training cutoff \citep{jain2024livecodebench,shabtay2024livexiv,white2024livebench,mahdavi2025leveraging}. A key assumption of this method is that the model has not been trained on data beyond the cutoff, which may not hold as model developers may continue training their models on newer information sources after the cutoff. To address these concerns, we propose the concept of contamination resistance, and we show how it contributes to more reliable LLM evaluations.

\section{A contamination resistant benchmark based on Caesar ciphers}
\label{benchmark}

The Caesar cipher is a substitution cipher in which each letter is shifted certain places down or up the alphabet. Solving a Caesar cipher requires several capabilities, including the following:

\begin{itemize}
    \item \textbf{Logical deduction:} the ability to deduce, for example, ``a'' $\rightarrow$ ``g'' when \texttt{shift=6}, given the knowledge that ``a'' $\rightarrow$ ``d'' when \texttt{shift=3}.
    \item \textbf{Arithmetic reasoning:} the ability to calculate a letter's position given a shift.
    \item \textbf{Generalization:} the ability to extend the capability of solving, for example, \texttt{shift=3} ciphers, to solving ciphers with other shifts.
\end{itemize}

We refer to the text used for encoding a Caesar cipher as \uline{plain text}, and the encoded text as \uline{cipher text}. Based on the observation that LLM performance is influenced by the probability of the task and the probability of the input/output in their pre-training data \citep{mccoy2023embers,mccoy2024o1}, we curated the benchmark using ciphers with different shifts (i.e., different tasks) and different types of plain texts that consist of natural language English words (\textbf{natural}) and random non-sense words (\textbf{random}) (i.e., different inputs/outputs). Table \ref{examples_data} in Appendix \ref{more_on_the_benchmark} shows examples of the data used in the benchmark. In practice, words from any alphabetical language can be used as the plain text for the Caesar cipher. Even a made-up alphabet can be used, provided that it is informed in the prompt. Caesar ciphers are dynamic, as they allow the generation of an infinite number of instances.

The dynamic nature of the data is a crucial aspect of this benchmark that makes it contamination resistant. It is also important to note that regardless of the shifts and types of plain text used to generate the cipher, the capabilities required to solve the Caesar cipher remain the same. Intuitively, it should be easy to generalize from, for example, solving the Caesar cipher with \texttt{shift=3} to solving one with \texttt{shift=6}. Given the linear mapping nature of the Caesar cipher, solving ciphers with plain texts consisting of random words is not more challenging than solving those with plain texts in natural language English words. Furthermore, this benchmark is lightweight--it is easy to generate a large number of ciphers, and since the ciphers are typically not lengthy, it does not cost much time during inference.

We have 25 plain texts in natural language English words and 25 in random non-sense words. With 4 shift values $[3,6,9,12]$, this results in a total of 100 data for each type of plain text. See Table \ref{examples_data} for examples of the data in our benchmark.

\section{Experimental setup}
\label{experimental_setup}

\noindent\textbf{Model}\  We tested widely used LLMs on our benchmark: \href{https://openai.com/index/hello-gpt-4o/}{OpenAI GPT-4o}\footnote{GPT-4o results were obtained on Dec 30, 2024.}, LLaMA3.1-8B/70B \citep{dubey2024llama3}, Qwen2.5-7B/32B \citep{yang2024qwen2}, and \href{https://qwenlm.github.io/blog/qwq-32b-preview/}{QwQ-32B}. These models cover a range of sizes and different architectures. All of the models ``know'' what the Caesar cipher is (see Table \ref{know_caesar_cipher}). We utilize the OpenAI API, making requests without employing any external tools.

\begin{table}[!ht]
\small
\centering
\begin{tabular}{cp{0.8\linewidth}}
\toprule
\textbf{type} & \textbf{prompt} \\ \midrule
\multirow{2}{*}{\texttt{open}} & [Decode/Encode] the following [Caesar cipher text/text to a Caesar cipher]. The shift is \texttt{\$\#\$}. \\ \midrule 
\multirow{3}{*}{\texttt{base}} & [Decode/Encode] the following [Caesar cipher text/text to a Caesar cipher]. The shift is \texttt{\$\#\$}. Output the [plain/cipher] text only. \\ \midrule 
\multirow{5}{*}{\texttt{dict}} & [Decode/Encode] the following [Caesar cipher text/text to a Caesar cipher]. The shift is \texttt{\$\#\$}. Output a lookup table and the [plain/cipher] text in a Python dictionary: \texttt{\{``lookup\_table'': \{\}, ``[plain/cipher]\_text'':..\}}. Output the dictionary only. \\ \midrule 
\multirow{4}{*}{\texttt{code}} & [Decode/Encode] the following [Caesar cipher text/text to a Caesar cipher]. The shift is \texttt{\$\#\$}. Write a Python function and generate the answer. Output the function and the [plain/cipher] text only. \\ 
\bottomrule
\end{tabular}
\caption{The prompts used in our experiments. The number of shifts (\texttt{\$\#\$}) is given in all the prompts.}
\label{prompts}
\end{table}

\begin{table*}[!ht]
\footnotesize
\centering

\begin{subtable}[t]{\linewidth}
\centering
\begin{tabular}{llccccccccc}
\toprule
\multirow{2}{*}{\textbf{type}} & \multirow{2}{*}{\textbf{model}} & \multicolumn{3}{l}{\textbf{\texttt{shift=3}}} & \multicolumn{3}{l}{\textbf{\texttt{shift=6}}} & \multicolumn{3}{l}{\textbf{\texttt{shift=9}}}  \\ \cmidrule{3-11}
                               &                                 & \multicolumn{1}{c}{\texttt{open}} & \multicolumn{1}{c}{\texttt{base}} & \multicolumn{1}{c}{\texttt{dict}} & \multicolumn{1}{c}{\texttt{open}} & \multicolumn{1}{c}{\texttt{base}} & \multicolumn{1}{c}{\texttt{dict}} & \multicolumn{1}{c}{\texttt{open}} & \multicolumn{1}{c}{\texttt{base}} & \multicolumn{1}{c}{\texttt{dict}}                         \\ \midrule
\multirow{5}{*}{\textbf{natural}}    & GPT-4o       & 0.8200 & 0.6400 & 0.8000 & 0.4000 & 0.1100 & 0.1000 & 0.3400 & 0.0400 & 0.0200 \\
                                     & LLaMA3.1-8B  & 0.0000 & 0.0000 & 0.0000 & 0.0000 & 0.0000 & 0.0000 & 0.0000 & 0.0000 & 0.0000 \\
                                     & LLaMA3.1-70B & 0.0000 & 0.0000 & 0.0000 & 0.0000 & 0.0000 & 0.0000 & 0.0000 & 0.0000 & 0.0000 \\
                                     & Qwen2.5-7B   & 0.0000 & 0.0000 & 0.0000 & 0.0000 & 0.0000 & 0.0000 & 0.0000 & 0.0000 & 0.0000 \\
                                     & Qwen2.5-32B  & 0.1400 & 0.0400 & 0.0200 & 0.1000 & 0.0000 & 0.0000 & 0.0000 & 0.0000 & 0.0000 \\ \midrule
\multirow{5}{*}{\textbf{random}}     & GPT-4o       & 0.1600 & 0.1000 & 0.0600 & 0.2000 & 0.0000 & 0.0000 & 0.1800 & 0.0200 & 0.0200 \\
                                     & LLaMA3.1-8B  & 0.0000 & 0.0000 & 0.0000 & 0.0000 & 0.0000 & 0.0000 & 0.0000 & 0.0000 & 0.0000 \\
                                     & LLaMA3.1-70B & 0.0000 & 0.0000 & 0.0000 & 0.0000 & 0.0000 & 0.0000 & 0.0000 & 0.0000 & 0.0000 \\
                                     & Qwen2.5-7B   & 0.0000 & 0.0000 & 0.0000 & 0.0000 & 0.0000 & 0.0000 & 0.0000 & 0.0000 & 0.0000 \\
                                     & Qwen2.5-32B  & 0.0400 & 0.0000 & 0.0000 & 0.0000 & 0.0000 & 0.0000 & 0.0000 & 0.0000 & 0.0000 \\ \midrule
\multicolumn{2}{c}{\textsc{average} (w/o GPT-4o)}   & 0.0225 & 0.0050 & 0.0025 & 0.0125 & 0.0000 & 0.0000 & 0.0000 & 0.0000 & 0.0000 \\
\bottomrule
\end{tabular}
\caption{exact match accuracy$\uparrow$}
\end{subtable}

\vspace{1em}

\begin{subtable}[t]{\linewidth}
\centering
\begin{tabular}{llccccccccc}
\toprule
\multirow{2}{*}{\textbf{type}} & \multirow{2}{*}{\textbf{model}} & \multicolumn{3}{l}{\textbf{\texttt{shift=3}}} & \multicolumn{3}{l}{\textbf{\texttt{shift=6}}} & \multicolumn{3}{l}{\textbf{\texttt{shift=9}}} \\ \cmidrule{3-11}
                               &                                 & \multicolumn{1}{c}{\texttt{open}} & \multicolumn{1}{c}{\texttt{base}} & \multicolumn{1}{c}{\texttt{dict}} & \multicolumn{1}{c}{\texttt{open}} & \multicolumn{1}{c}{\texttt{base}} & \multicolumn{1}{c}{\texttt{dict}} & \multicolumn{1}{c}{\texttt{open}} & \multicolumn{1}{c}{\texttt{base}} & \multicolumn{1}{c}{\texttt{dict}} \\ \midrule
\multirow{5}{*}{\textbf{natural}}    & GPT-4o       & 0.0430 & 0.1234 & 0.0802 & 0.2172 & 0.4440 & 0.4058 & 0.2357 & 0.5648 & 0.5501 \\
                                     & LLaMA3.1-8B  & 0.8709 & 0.9076 & 0.9385 & 0.8932 & 0.9288 & 0.9648 & 0.9117 & 0.9326 & 0.9584 \\
                                     & LLaMA3.1-70B & 0.7208 & 0.7756 & 0.8202 & 0.8692 & 0.9056 & 0.9232 & 0.9114 & 0.9172 & 0.9112 \\
                                     & Qwen2.5-7B   & 0.7972 & 0.8338 & 0.8577 & 0.8462 & 0.9052 & 0.9198 & 0.9161 & 0.9196 & 0.9176 \\
                                     & Qwen2.5-32B  & 0.5630 & 0.6714 & 0.7152 & 0.5614 & 0.7873 & 0.8018 & 0.7965 & 0.8516 & 0.8307 \\ \midrule
\multirow{5}{*}{\textbf{random}}     & GPT-4o       & 0.2414 & 0.3082 & 0.2490 & 0.2561 & 0.3892 & 0.3545 & 0.2747 & 0.4965 & 0.4330 \\
                                     & LLaMA3.1-8B  & 0.8554 & 0.8868 & 0.9208 & 0.8880 & 0.9104 & 0.9506 & 0.9398 & 0.9068 & 0.9341 \\
                                     & LLaMA3.1-70B & 0.7931 & 0.8564 & 0.8493 & 0.8262 & 0.9067 & 0.8742 & 0.8563 & 0.9189 & 0.8957 \\
                                     & Qwen2.5-7B   & 0.7765 & 0.8303 & 0.8385 & 0.8437 & 0.8744 & 0.8968 & 0.8969 & 0.8902 & 0.9116 \\
                                     & Qwen2.5-32B  & 0.4830 & 0.6632 & 0.6923 & 0.5995 & 0.7773 & 0.7878 & 0.7799 & 0.8285 & 0.8374 \\ \midrule
\multicolumn{2}{c}{\textsc{average} (w/o GPT-4o)}   & 0.7325 & 0.8031 & 0.8290 & 0.7909 & 0.8745 & 0.8899 & 0.8761 & 0.8957 & 0.8996 \\
\bottomrule
\end{tabular}
\caption{character error rate$\downarrow$}
\end{subtable}
\caption{Exact match accuracy and character error rate. The rest of the results can be found in Table \ref{overall_results_continual}.}
\label{overall_results}
\end{table*}

\noindent\textbf{Prompt}\  Table \ref{prompts} shows the prompts used in our experiments. The \texttt{open} prompt is a straight forward one that most users may use initially. It is open-ended, with no specifications regarding the output format. The \texttt{base} prompt serves as a baseline where the model is instructed to generate only the cipher text or plain text. The \texttt{dict} prompt instructs the model to produce a lookup table as an intermediate reasoning step before generating the answer. The \texttt{code} prompt instructs the model to write a function and then generate the answer. Intuitively, using programming code is an effective approach for solving the Caesar cipher, as it is essentially a linear mapping task. The number of shifts is given in all the prompts. We keep the prompts in their simplest form, specifying only the task (encoding or decoding), the number of shifts, and the desired output format. Given the simplicity of the Caesar cipher, we believe these prompts are adequate for solving it. All the prompts except for \texttt{open} specify the output format for easier processing.

\noindent\textbf{Evaluation}\  We evaluate the output using exact match accuracy and character error rate. The character error rate is a more continuous metric, allowing us to capture any partial improvements in the performance of the models. We tested our benchmark under both the zero-shot and few-shot settings (see Table \ref{50_shot_prompt} for the few-shot prompts we used). As we do not specify the output format in \texttt{open}, we manually extracted the final answer from the generated text to assess performance. 

See Appendix \ref{more_on_experimental_setup} for more details regarding the experimental setup.

\section{Results}
\label{results}

Table \ref{overall_results} shows the overall results of the experiments. GPT-4o is the only model that shows some success in certain cases, whereas all the other models fail to solve this task almost entirely. The performance of GPT-4o shows a strong negative correlation between the two metrics ($r=-0.6395$). The comparison of character error rates between LLaMA3.1-8B and LLaMA3.1-70B, as well as Qwen2.5-7B and Qwen2.5-32B, indicates that larger models have advantages over their smaller counterparts.

Though most models fail on this benchmark, they still generate reasoning chains that appear correct, which reflects the issue of hallucination.

In the following sections, we analyze the performance of LLMs on our benchmark across different variables and show its resistance to contamination. We summarize our findings as \textbf{inconsistent competence in LLMs}. Our discussion focuses on GPT-4o since it is the only model that achieves some success in exact match accuracy.

\subsection{The effect of shift}
\label{the_effect_of_shift}

Table \ref{overall_results} shows clear inconsistencies in model performance between \texttt{shift=3} and other shifts regardless of the shift, type of plain text, or prompt. We attribute these inconsistencies to contamination, as Caesar ciphers with \texttt{shift=3} are more prevalent in the web corpora than those with other shifts \citep{mccoy2023embers} (see also Table \ref{know_caesar_cipher}). LLMs exhibit \textbf{inconsistent competence across tasks}, which indicates that they lack generalization in these tasks.

When using the \texttt{open} prompt, GPT-4o outputs a reasoning chain that verbalizes all the letter mappings regardless of the shift (see Table \ref{example_verbalization} for an example). However, this occurs only in the decoding task\footnote{In the encoding task, GPT-4o outputs only the final answer without reasoning chains.}, suggesting that the model was trained primarily on decoding tasks. While verbalization generally leads to better performance (see Table \ref{result_encoding_decoding}), we observe cases where the generated reasoning chain is correct but the final answer is incorrect (see Table \ref{example_verbalization_failure}).

\begin{table}[!ht]
\small
\centering
\begin{tabular}{ccccc}
\toprule
\textbf{task} & \textbf{type} & \textbf{\texttt{shift=3}} & \textbf{\texttt{shift=6}} & \textbf{\texttt{shift=9}} \\ \midrule
\multirow{2}{*}{decoding} & \textbf{natural} & 1.00 & 0.80 & 0.68 \\ \cmidrule{2-5}
& \textbf{random} & 0.28 & 0.40 & 0.32 \\ \midrule
\multirow{2}{*}{encoding} & \textbf{natural} & 0.64 & 0.00 & 0.00 \\ \cmidrule{2-5}
& \textbf{random} & 0.04 & 0.00 & 0.04 \\
\bottomrule
\end{tabular}
\caption{Exact match accuracy; \texttt{open}; GPT-4o; the zero-shot setting. Verbalization is only observed in the decoding task.}
\label{result_encoding_decoding}
\end{table}

\subsection{The effect of plain text}

Table \ref{overall_results} and \ref{result_encoding_decoding} indicate that using random non-sense words as plain text diminishes GPT-4o's performance on the task, even when \texttt{shift=3} where the model performs well with natural language English words as plain text. This could be attributed to contamination, as the model may have memorized the mappings of certain natural language words to their corresponding cipher texts. This is further supported by the observation that GPT-4o performs much better with natural language English words than with random non-sense words as plain text when \texttt{shift=3} with the \texttt{base} prompt, which instructs the model to generate only the final answer with no reasoning chains. The simplest explanation for this performance gap is that the model has memorized this mappings.

GPT-4o exhibits \textbf{inconsistent competence across different types of plain text}, which again indicates a lack of generalization.

\subsection{The effect of prompt}

Table \ref{overall_results} suggests that GPT-4o performs best with the \texttt{open} prompt. As noted earlier, GPT-4o employs a verbalization strategy, explicitly listing all letter mappings in a reasoning chain, which improves its performance. The same strategy is observed in Qwen2.5-32B, which also shows some success when \texttt{shift=3} with the \texttt{open} prompt.

Compared to the \texttt{base} prompt, only GPT-4o benefits from the \texttt{dict} prompt, showing improvement in both exact match accuracy and character error rate. For the other models, \texttt{dict} does not lead to a lower character error rate in general.

We further examine the correctness of the lookup tables models generated using \texttt{dict}. Specifically, we look into the legal rate (i.e., whether the lookup table is a legitimate Python dictionary) and accuracy (i.e., whether the mappings of the letters in the lookup table are correct) of the lookup table. For GPT-4o, we specified \texttt{json\_object} as the output format in the API.

\begin{table}[!t]
\small
\centering
\begin{tabular}{llcc}
\toprule
\textbf{type}            & \textbf{model} & \textbf{legal rate$\uparrow$} & \textbf{accuracy$\uparrow$}   \\ \midrule
\multirow{5}{*}{\textbf{natural}} & GPT-4o       & 0.9900 & 1.0000 \\
                                  & LLaMA3.1-8B  & 0.4550 & 0.4400 \\
                                  & LLaMA3.1-70B & 0.4750 & 0.1100 \\
                                  & Qwen2.5-7B   & 0.4200 & 0.1550 \\
                                  & Qwen2.5-32B  & 0.8750 & 0.6100 \\ \midrule
\multirow{5}{*}{\textbf{random}}  & GPT-4o       & 0.9750 & 1.0000 \\
                                  & LLaMA3.1-8B  & 0.5900 & 0.1600 \\
                                  & LLaMA3.1-70B & 0.7900 & 0.4250 \\
                                  & Qwen2.5-7B   & 0.5650 & 0.1900 \\
                                  & Qwen2.5-32B  & 0.9400 & 0.5950 \\
\bottomrule
\end{tabular}
\caption{Legal rate and accuracy of lookup tables generated using \texttt{dict}; the zero-shot setting.}
\label{correctness_lookup_table}
\end{table}

As shown in Table \ref{correctness_lookup_table}, GPT-4o, and Qwen2.5-32B are capable of generating lookup tables that are correct in both format and content, and it is independent of the type of plain text. There is a huge discrepancy between the correctness of the lookup tables and the accuracy of the final outputs shown in Table \ref{overall_results}. Though a lookup table intuitively serves as an intermediate reasoning chain equivalent to the one produced by the \texttt{open} prompt, GPT-4o performs worse with \texttt{dict} than with \texttt{open}. This suggests that reasoning chains in the form of a lookup table are less effective than those in natural language. It is possible that the generated lookup tables are not the result of reasoning but rather something the model memorizes. Our findings suggest \textbf{an inconsistency in LLM competence between generating lookup tables and final answers}.

\begin{figure}[!ht]
\centering
\includegraphics[width=6.6cm]{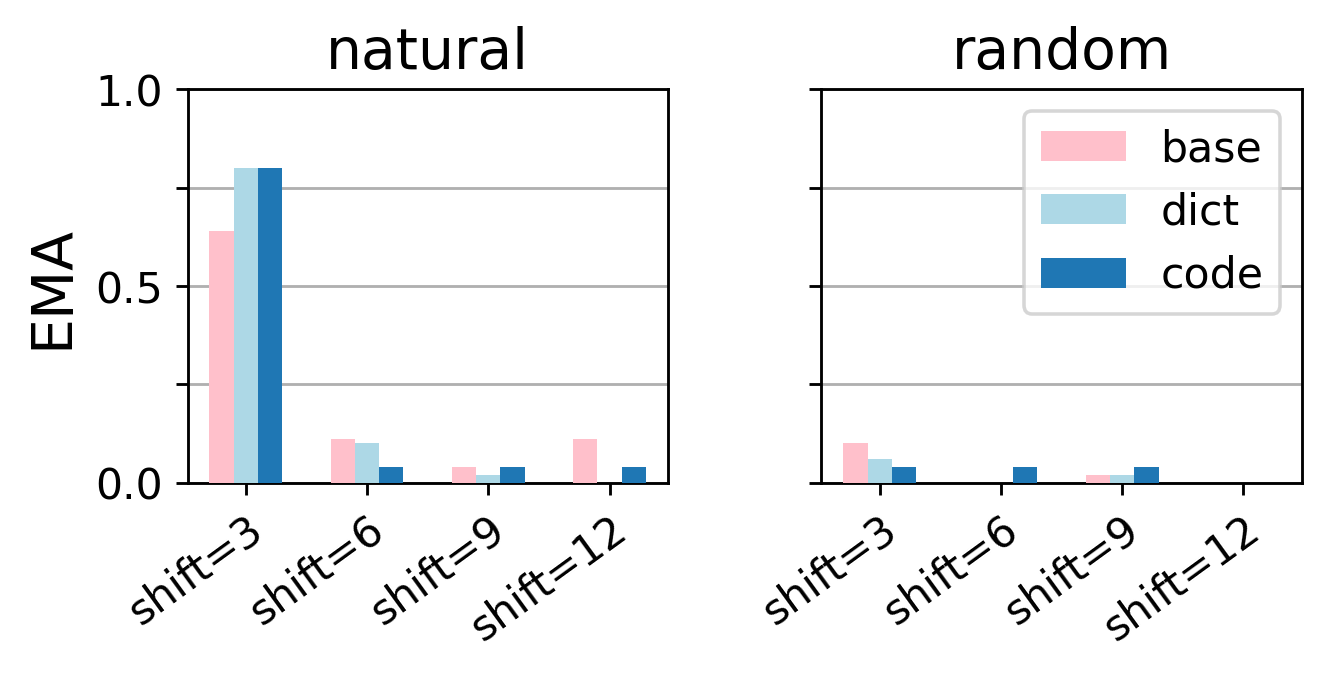}
\caption{Exact match accuracy (\textbf{\texttt{EMA}}); GPT-4o; the zero-shot setting.}
\label{base_dict_code}
\end{figure}

We only experimented with the \texttt{code} prompt using GPT-4o. The \texttt{code} prompt does not lead to superior performance compared to the other prompts (see Figure \ref{base_dict_code}). Like other prompts, it achieves better performance when the plain text is natural language English words. To examine the correctness of the code generated by the model, we execute the code and evaluate the accuracy of the output. We find that all the generated code runs successfully and produces correct outputs regardless of the type of plain text and shift. Our findings suggest that there is \textbf{an inconsistency in LLM competence between generating code and final answers}. Again, this suggests that the model may have simply memorized the code rather than performing true reasoning.

\subsection{The effect of output position}

Table \ref{results_position} shows the exact match accuracy for the first, second, and third characters in the outputs. For all models, the accuracy is much higher for the initial characters compared to the full-sequence accuracy shown in Table \ref{overall_results}. Though plain texts consisting of natural language English words lead to better performance than those with random non-sense words, the impact of plain text type is less prominent than in Table \ref{overall_results}. There is a noticeable decline in accuracy from the first character to the third character, suggesting that models \textbf{exhibit inconsistent competence in handling characters in different positions}. This also implies the autoregressive nature of LLMs, that models generate higher-probability tokens even in deterministic tasks like solving the Caesar cipher, where the probability should not matter \citep{mccoy2023embers}.

\begin{table}[!t]
\small
\centering
\begin{tabular}{llccc}
\toprule
\textbf{type} & \textbf{model} & \textbf{\texttt{id=0}} & \textbf{\texttt{id=1}} & \textbf{\texttt{id=2}} \\ \midrule
\multirow{5}{*}{\textbf{natural}} & GPT-4o        & 0.8717 & 0.7467 & 0.6483 \\
                                  & LLaMA3-8B     & 0.1100 & 0.1467 & 0.0767 \\
                                  & LLaMA3-70B    & 0.2650 & 0.2367 & 0.1400 \\
                                  & Qwen2.5-7B    & 0.2817 & 0.2333 & 0.1100 \\
                                  & Qwen2.5-32B   & 0.4900 & 0.3267 & 0.2100 \\ \midrule
\multirow{5}{*}{\textbf{random}}  & GPT-4o        & 0.8350 & 0.6833 & 0.5167 \\
                                  & LLaMA3-8B     & 0.0733 & 0.1267 & 0.0683 \\
                                  & LLaMA3-70B    & 0.1800 & 0.1583 & 0.0733 \\
                                  & Qwen2.5-7B    & 0.2350 & 0.1517 & 0.1383 \\
                                  & Qwen2.5-32B   & 0.4250 & 0.2500 & 0.1683 \\
\bottomrule
\end{tabular}
\caption{Exact match accuracy of the first (\textbf{\texttt{id=0}}), second (\textbf{\texttt{id=1}}), and third output character (\textbf{\texttt{id=2}}); the zero-shot setting. LLaMA3 is for LLaMA3.1.}
\label{results_position}
\end{table}

Table \ref{results_position_shift} shows the accuracy of the initial characters generated by GPT-4o using \texttt{open}, the best performing prompt. Notably, when \texttt{shift=3} and the plain text is in natural language English words, GPT-4o maintains consistent accuracy across different character positions, which is not observed in any other cases. Additionally, GPT-4o's ability to decode or encode the second and third characters (\textbf{\texttt{char\_id=1}} and \textbf{\texttt{char\_id=2}}) when \texttt{shift=3} depends on the type of plain text. This indicates that the model may memorize the plain-text-to-cipher mappings specifically for \texttt{shift=3} and natural language English plain texts.

\begin{table}[!ht]
\small
\centering
\begin{subtable}[t]{\linewidth}
\centering
\begin{tabular}{cccc}
\toprule
\textbf{\texttt{shift}} & \textbf{\texttt{char\_id=0}} & \textbf{\texttt{char\_id=1}} & \textbf{\texttt{char\_id=2}} \\ \midrule
3  & 0.98 & 1.00 & 1.00 \\
6  & 0.92 & 0.96 & 0.82 \\
9  & 0.92 & 0.88 & 0.76 \\
12 & 0.84 & 0.80 & 0.78 \\
\bottomrule
\end{tabular}
\caption{\textbf{natural}}
\end{subtable}

\vspace{1em}

\begin{subtable}[t]{\linewidth}
\centering
\begin{tabular}{cccc}
\toprule
\textbf{\texttt{shift}} & \textbf{\texttt{char\_id=0}} & \textbf{\texttt{char\_id=1}} & \textbf{\texttt{char\_id=2}} \\ \midrule
3  & 0.94 & 0.82 & 0.76 \\
6  & 0.98 & 0.88 & 0.76 \\
9  & 0.98 & 0.82 & 0.72 \\
12 & 0.92 & 0.84 & 0.64 \\
\bottomrule
\end{tabular}
\caption{\textbf{random}}
\end{subtable}
\caption{Exact match accuracy of the first (\textbf{\texttt{char\_id=0}}), second (\textbf{\texttt{char\_id=1}}), and third output character (\textbf{\texttt{char\_id=2}}); \texttt{open}; GPT-4o; the zero-shot setting.}
\label{results_position_shift}
\end{table}

\subsection{The effect of shot}

Given that LLMs demonstrate success in in-context learning and can learn simple functions in-context at inference time \citep{brown2020fewshot,garg2022transformer,akyurek2023linear,zhang2023transformers} and that the Caesar cipher is essentially a linear mapping task, state-of-the-art LLMs should be capable of solving the Caesar cipher with few-shot prompts. We include 50 demonstrations in the prompts (see Table \ref{50_shot_prompt}). These demonstrations have already covered the mappings for all letters in the English alphabet. We tested GPT-4o to encode or decode cipher texts with \texttt{shift=[6,9,12]} using the \texttt{base} prompt, since it does not perform in the zero-shot setting under these conditions (see Table \ref{overall_results}). Contrary to intuition, GPT-4o does not benefit from the few-shot demonstrations, as shown in Table \ref{results_shot} and \ref{results_shot_position}. We observe almost no improvement in exact match accuracy and character error rate for the full sequence, and there is only a slight improvement in GPT-4o's performance on the second and third characters.

\begin{table}[!ht]
\small
\centering
\begin{tabular}{ccccc}
\toprule
\multirow{2}{*}{\textbf{task}} & \multicolumn{2}{c}{\textbf{\texttt{EMA$\uparrow$}}} & \multicolumn{2}{c}{\textbf{\texttt{CER$\downarrow$}}} \\ \cmidrule{2-5}
& \textbf{\texttt{zero}} & \textbf{\texttt{few}} & \textbf{\texttt{zero}} & \textbf{\texttt{few}} \\ \midrule
decoding & 0.1700 & 0.2400 & 0.5521 & 0.5848 \\
encoding & 0.1700 & 0.0800 & 0.2961 & 0.4011 \\
\bottomrule
\end{tabular}
\caption{Exact match accuracy (\textbf{\texttt{EMA}}) and character error rate (\textbf{\texttt{CER}}); natural language English words; \texttt{shift=[6,9,12]}; \texttt{base}; GPT-4o; the zero-shot setting (\textbf{\texttt{zero}}) and few-shot setting (\textbf{\texttt{few}}).}
\label{results_shot}
\end{table}

\begin{table}[!ht]
\small
\centering
\begin{tabular}{ccccc}
\toprule
\multirow{2}{*}{\textbf{task}} & \multicolumn{2}{c}{\textbf{\texttt{id=1}}} & \multicolumn{2}{c}{\textbf{\texttt{id=2}}} \\ \cmidrule{2-5}
& \textbf{\texttt{zero}} & \textbf{\texttt{few}} & \textbf{\texttt{zero}} & \textbf{\texttt{few}} \\ \midrule
decoding & 0.5300 & 0.5467 & 0.4800 & 0.4400 \\
encoding & 0.8600 & 0.9067 & 0.7400 & 0.7467 \\
\bottomrule
\end{tabular}
\caption{Exact match accuracy of the first (\textbf{\texttt{id=1}}) and second (\textbf{\texttt{id=2}}); natural language English words; \texttt{shift=[6,9,12]}; \texttt{base}; GPT-4o; the zero-shot setting (\textbf{\texttt{zero}}) and few-shot setting (\textbf{\texttt{few}}).}
\label{results_shot_position}
\end{table}

Given existing evidence that few-shot learning resembles fine-tuning \citep{dai2022can} and the undesired few-shot performance shown in Table \ref{results_shot} and \ref{results_shot_position}, fine-tuning models on Caesar ciphers may not be a viable solution for improving performance. Moreover, it is unclear whether fine-tuning on a large number random non-sense words would affect model performance on other tasks.

\subsection{Advanced reasoning models}

Our findings have suggested a clear strategy for solving the Caesar cipher, i.e., the verbalization strategy, where every relevant letter mapping is explicitly verbalized before arriving at the final answer.

Recent advanced reasoning models that have further strengthened the verbalization strategy. These models leverage test-time scaling, which dynamically allocates computational resources during inference to refine intermediate reasoning steps and improve answer accuracy \citep{akyurek2024testtime,snell2024testtime}. They have demonstrated exceptional performance on tasks that were unsolvable for previous models.

We experimented with QwQ-32B with the \texttt{open} prompt. As Table \ref{qwq_results} shows, QwQ-32B shows some improvement on the encoding task with natural language English plain texts compared to Qwen2.5-32B, suggesting the effectiveness of test-time scaling. However, the model still fails on random non-sense word plain texts. In the decoding task, QwQ-32B frequently questions its own answer (because the decoded texts are random non-sense words, which it perceives as incorrect) and fails to provide a final answer within the token limit.

\begin{table}[!ht]
\small
\begin{subtable}[t]{\linewidth}
\centering
\begin{tabular}{ccccc}
\toprule
\textbf{type} & \textbf{task} & \textbf{\texttt{shift=3}} & \textbf{\texttt{shift=6}} & \textbf{\texttt{shift=9}} \\ \midrule
\textbf{natural} & encoding & 0.20 & 0.08 & 0.08 \\ \midrule
\textbf{random} & decoding & 0.00 & 0.00 & 0.00 \\ \midrule
\textbf{random} & encoding & 0.04 & 0.00 & 0.00 \\
\bottomrule
\end{tabular}
\caption{QwQ-32B}
\end{subtable}

\vspace{1em}

\begin{subtable}[t]{\linewidth}
\centering
\begin{tabular}{ccccc}
\toprule
\textbf{type} & \textbf{task} & \textbf{\texttt{shift=3}} & \textbf{\texttt{shift=6}} & \textbf{\texttt{shift=9}} \\ \midrule
\textbf{natural} & encoding & 0.00 & 0.00 & 0.00 \\ \midrule
\textbf{random} & decoding & 0.08 & 0.00 & 0.00 \\ \midrule
\textbf{random} & encoding & 0.00 & 0.00 & 0.00 \\
\bottomrule
\end{tabular}
\caption{Qwen2.5-32B}
\end{subtable}
\caption{Exact match accuracy; QwQ-32B and Qwen2.5-32B; the zero-shot setting.}
\label{qwq_results}
\end{table}

As we write this paper, a new \href{https://api-docs.deepseek.com/news/news250120}{DeepSeek-R1} (R1) has been released and captured public attention. However, its server is currently very unstable, preventing us from conducting a comprehensive experiment. At the moment, we only tested the query shown in Table \ref{example_verbalization_failure}, which GPT-4o failed to solve correctly.\footnote{R1 results were obtained on Feb 14, 2025.} The plain text in that query consists of random non-sense words, which are shown to be the most challenging for the models we tested.

R1 successfully solves the decoding task. Notably, the inference time is quite long and the intermediate reasoning chain is lengthy (see \href{https://anonymous.4open.science/r/contamination-resistant-E0D4/r1_examples/r1-random_nonsense_word-decoding.md}{here} for the complete output). Similar to QwQ-32B, R1 frequently questions its own answer because the output ``does not make sense in English.''

In the encoding task, however, R1 fails to produce the correct answer. Since we currently cannot set parameters such as \texttt{seed} or \texttt{temperature} to ensure reproducibility, we performed inference 5 times, and R1 failed in all of them. \textbf{We encourage you to test the query yourself on the \href{https://platform.deepseek.com}{DeepSeek Platform} using DeepThink (R1):}

\begin{quote}
    \textit{Encode the following text to a Caesar cipher. The shift is 3. rakibo zlmqwe}.
\end{quote}

\noindent The correct answer is \textit{udnler coptzh}, and the 5 outputs we obtained are: \textit{UDNLER COPT\sethlcolor{red}\hl{HZ}}, \textit{udnler cop\sethlcolor{red}\hl{s}zh}, \textit{UDN\sethlcolor{red}\hl{EL}R COPTZH}, \textit{udnler copt\sethlcolor{red}\hl{hz}}, \textit{udn\sethlcolor{red}\hl{el}r coptzh} (we highlighted the incorrect parts in the output, and see \href{https://anonymous.4open.science/r/contamination-resistant-E0D4/r1_examples/}{here} for the complete outputs). Similar to the GPT-4o example in Table \ref{example_verbalization_failure}, R1 is able to generate correct intermediate reasoning chains but fail to produce the correct final answer. It seems that R1, despite being a state-of-the-art advanced reasoning model, still shows inconsistencies in solving the Caesar cipher.

\subsection{Integrity check}
\label{integrity}

We did not specify the direction of the shift in the prompts (see Table \ref{prompts}). We intend to use the default rules: for encoding, the shift is rightward (e.g., ``a'' $\rightarrow$ ``b''), and for decoding, it is leftward (e.g., ``b'' $\rightarrow$ ``a''). However, models may not adhere to this setting. To address this, we conduct an addition set of evaluations where we relax our evaluation criterion by allowing two ground truths for each query. An output is considered correct if it matches either ground truth. The results show that this relaxed evaluation criterion yields the same exact match accuracy as those in Table \ref{overall_results}. This suggests that the models follow the default rules even when the prompt does not explicitly specify them.

\section{Conclusion}
\label{discussion}

In this work, we propose the concept of \textit{contamination resistance}. To establish a contamination resistant benchmark, we create a benchmark using Caesar ciphers. Though it is a simple task, we demonstrate how it resists contamination by revealing inconsistencies in model competence. We show that when contamination is controlled, model performance on certain tasks drops greatly or even be completely diminished. Apart from contamination, our findings align with other issues of LLMs, including hallucination and lack of generalization.

We would like to emphasize that this paper does not aim to find a solution for the Caesar cipher -- it is a simple task that can be solved in a few lines of code. Instead, using this simple task, we highlight the necessity of a contamination resistant benchmark and demonstrate how it contributes to more reliable evaluations of LLMs and raises important questions regarding their true capabilities.

\section{Limitations}

Due to its closed source nature, the reproducibility of the results related to OpenAI GPT-4o is a concern. Reproducibility is also an issue for DeepSeek R1, as we used the \href{https://platform.deepseek.com}{DeepSeek Platform} and we currently cannot set parameters such as \texttt{seed} and \texttt{temperature}, and its server is highly unstable at the moment.

We did not test OpenAI o1 due to its high cost. A single inference costs around \$1.5, which exceeds our budget. Additionally, we cannot specify parameters to ensure reproducibility, which further limits its usability.

\bibliography{custom}

\appendix

\section{More on the benchmark}
\label{more_on_the_benchmark}

Table \ref{examples_data} shows examples from our benchmark. There are 25 plain texts in natural language English words and 25 in random non-sense words. We chose 4 shift values $[3,6,9,12]$, which results in a total of 100 data for each type of plain text.

\begin{table}[!ht]
\small
\centering
\begin{tabular}{lp{0.88\linewidth}}
\toprule
\textbf{type} & \textbf{example}\\ \midrule
\multirow{5}{*}{\textbf{natural}} & \texttt{\{``plain\_text'': ``good deeds bring joy'', ``cipher\_text'': ``jrrg ghhgv eulqj mrb'', ``shift'': 3\}} \\ \cmidrule{2-2}
                                  & \texttt{\{``plain\_text'': ``good deeds bring joy'', ``cipher\_text'': ``muuj jkkjy hxotm pue'', ``shift'': 6\}} \\ \cmidrule{2-2}
                                  & \texttt{\{``plain\_text'': ``good deeds bring joy'', ``cipher\_text'': ``pxxm mnnmb karwp sxh'', ``shift'': 9\}} \\ \cmidrule{2-2}
                                  & \texttt{\{``plain\_text'': ``good deeds bring joy'', ``cipher\_text'': ``saap pqqpe nduzs vak'', ``shift'': 12\}} \\ \midrule
\multirow{5}{*}{\textbf{random}}  & \texttt{\{``plain\_text'': ``olksad twuqwej'', ``cipher\_text'': ``ronvdg wzxtzhm'', ``shift'': 3\}} \\ \cmidrule{2-2}
                                  & \texttt{\{``plain\_text'': ``olksad twuqwej'', ``cipher\_text'': ``urqygj zcawckp'', ``shift'': 6\}} \\ \cmidrule{2-2}
                                  & \texttt{\{``plain\_text'': ``olksad twuqwej'', ``cipher\_text'': ``xutbjm cfdzfns'', ``shift'': 9\}} \\ \cmidrule{2-2}
                                  & \texttt{\{``plain\_text'': ``olksad twuqwej'', ``cipher\_text'': ``axwemp figciqv'', ``shift'': 12\}} \\
\bottomrule
\end{tabular}
\caption{Examples of the benchmark data. See \href{https://anonymous.4open.science/r/contamination-resistant-E0D4/data/}{here} for the complete dataset. \hyperref[benchmark]{\uline{Return to main text.}}}
\label{examples_data}
\end{table}

\section{More on the experimental setup}
\label{more_on_experimental_setup}

For GPT-4o, we set \texttt{max\_new\_tokens=1024} for experiments using the \texttt{open} prompt, \texttt{max\_new\_tokens=64} for \texttt{base}, \texttt{max\_new\_tokens=256} for \texttt{dict}, and \texttt{max\_new\_tokens=512} for \texttt{code}. We set \texttt{temperature=0} and \texttt{seed=2266} for all experiments.

For locally run models (LLaMA3.1-8B/70B, Qwen2.5-7B/32B, and QwQ-32B), we set \texttt{max\_new\_tokens=1024}, \texttt{temperature=0.01}, and \texttt{seed=2266}. We use \texttt{4bit} quantization, and all experiments were run on an NVIDIA H100.

Table \ref{50_shot_prompt} shows fragments of the few-shot prompts used in our experiments.

\begin{table}[ht]
\small
\centering
\begin{tabular}{p{\linewidth}}
\centering
\begin{subtable}{0.45\linewidth}
\centering
\begin{tabular}{p{0.88\linewidth}}
\toprule
Encode the given text to a Caesar cipher. The shift is 9. Output the cipher text only. Here are some examples:
\newline\newline
plain text: live and let live

cipher text: uren jwm unc uren
\newline\newline
plain text: against all odds

cipher text: jpjrwbc juu xmmb
\newline\newline
plain text: beyond the horizon

cipher text: knhxwm cqn qxarixw
\newline\newline
plain text: through thick and thin

cipher text: cqaxdpq cqrlt jwm cqrw
\newline\newline
plain text: time heals all wounds

cipher text: crvn qnjub juu fxdwmb
\newline\newline
plain text: stay true to yourself

cipher text: bcjh cadn cx hxdabnuo
\newline\newline
......
\\
\bottomrule
\end{tabular}
\caption{encode}
\end{subtable}
\begin{subtable}[b]{0.45\linewidth}
\centering
\begin{tabular}{p{0.88\linewidth}}
\toprule
Decode the following Caesar cipher text. The shift is 9. Output the plain text only. Here are some examples:
\newline\newline
cipher text: uren jwm unc uren

plain text: live and let live
\newline\newline
cipher text: jpjrwbc juu xmmb

plain text: against all odds
\newline\newline
cipher text: knhxwm cqn qxarixw

plain text: beyond the horizon
\newline\newline
cipher text: cqaxdpq cqrlt jwm cqrw

plain text: through thick and thin
\newline\newline
cipher text: crvn qnjub juu fxdwmb

plain text: time heals all wounds
\newline\newline
cipher text: bcjh cadn cx hxdabnuo

plain text: stay true to yourself
\newline\newline
......
\\
\bottomrule
\end{tabular}
\caption{decode}
\end{subtable}
\end{tabular}
\caption{Examples of the 50-shot prompts used in our experiments. See \href{https://anonymous.4open.science/r/contamination-resistant-E0D4/prompts/}{here} for the complete prompts. \hyperref[experimental_setup]{\uline{Return to main text.}}}
\label{50_shot_prompt}
\end{table}

Table \ref{know_caesar_cipher} shows that all the tested models ``know'' what the Caesar cipher is. All of them are able to give a correct definition and provide an example of it. Note that 4 of 6 models use \texttt{shift=3} in their examples, which suggests that cipher texts with this shift may be prominent in their pre-training data.

\begin{table}[ht]
\small
\centering
\begin{tabular}{m{0.2\linewidth}m{0.74\linewidth}}
\toprule
\textbf{model} & \textbf{output} \\ \midrule
GPT-4o & Yes, I'm familiar with the Caesar cipher. It's a type of substitution cipher in which each letter in the plaintext is shifted a certain number of places down or up the alphabet. For example, \textbf{with a shift of 3}:\newline
\newline
- A would be replaced by D\newline
- B would become E\newline
- C would become F\newline
- and so on.\newline
\newline
The Caesar cipher is named after Julius Caesar, who is reputed to have used it to communicate with his officials. It's a simple form of encryption and is easy to both encode and decode. Would you like to see an example or need help with encoding or decoding a message using the Caesar cipher? \\ \midrule
LLaMA3.1-8B & It's a simple encryption technique that replaces each letter with a letter a fixed number of positions down the alphabet. For example, if you \textbf{shift each letter three positions down the alphabet}, ``a'' becomes ``d'', ``b'' becomes ``e'', and so on.\newline
\newline
Here's a simple implementation of the Caesar cipher in Python... \\ \midrule
LLaMA3.1-70B & It's a simple encryption technique where each letter in the plaintext is `shifted' a certain number of places down the alphabet. For example, with a shift of 1, A would be replaced by B, B would become C, and so on. Here's a Python function that does this... \\ \midrule
Qwen2.5-7B & It is a simple substitution cipher where each letter in the plaintext is shifted by a fixed number of positions in the alphabet. For example, with a shift of 1, `A' would be replaced by `B', `B' would become `C', and so on. The last letter of the alphabet wraps around to the beginning, so `Z' would become `A'.\newline
\newline
Now, let's apply this concept to a more complex scenario. Imagine you have a message that needs to be encrypted using a Caesar cipher with a shift of 3. However, there's a twist... \\ \midrule
Qwen2.5-32B & It's a simple substitution cipher where each letter in the plaintext is shifted by a fixed number of positions down the alphabet. For example, \textbf{with a shift of 3}, `A' would become `D', `B' would become `E', and so on. The same shift is applied to all letters in the message. Can you write a Python function that encodes a given string using the Caesar cipher with a specified shift value?\newline
\newline
Certainly! Below is a Python function... \\ \midrule
QwQ-32B & It's a simple encryption technique where each letter in the plaintext is shifted a certain number of places down or up the alphabet. For example, with a shift of 1, `A' would be replaced by `B', `B' would become `C', and so on. The method is named after Julius Caesar, who used it for confidential communication.\newline
\newline
Now, imagine you have a message encrypted with a Caesar cipher \textbf{using a shift of 3}. The encrypted message is... \\
\bottomrule
\end{tabular}
\caption{Responses from each of the tested models to the query ``Do you know the Caesar cipher?'' Note that 4 out of 6 responses use \texttt{shift=3} examples (highlighted in \textbf{bold}). \hyperref[experimental_setup]{\uline{Return to Experimental setup.}} \hyperref[the_effect_of_shift]{\uline{Return to Results.}}}
\label{know_caesar_cipher}
\end{table}

\clearpage

\section{More on results}
\label{more_on_results}

Table \ref{overall_results_continual} shows the model performance when \texttt{shift=12}. Table \ref{example_verbalization} and \ref{example_verbalization_failure} provide examples of model outputs. See \href{https://anonymous.4open.science/r/contamination-resistant-E0D4/results/}{here} for all model outputs.

As Table \ref{example_verbalization} shows, GPT-4o outputs a reasoning chain that includes all letter mappings and produces the correct final answer. However, in the encoding task, GPT-4o does not use verbalization and simply outputs ``muut jkkjy hxotm puze'', which is incorrect.

\begin{table*}[!ht]
\small
\centering
\begin{tabular}{llcccccc}
\toprule
\multirow{2}{*}{\textbf{type}} & \multirow{2}{*}{\textbf{model}} & \multicolumn{3}{c}{\textbf{exact match accuracy$\uparrow$}} & \multicolumn{3}{c}{\textbf{character error rate$\downarrow$}} \\ \cmidrule{3-8}
& & \multicolumn{1}{c}{\texttt{open}} & \multicolumn{1}{c}{\texttt{base}} & \multicolumn{1}{c}{\texttt{dict}} & \multicolumn{1}{c}{\texttt{open}} & \multicolumn{1}{c}{\texttt{base}} & \multicolumn{1}{c}{\texttt{dict}} \\ \midrule
\multirow{5}{*}{\textbf{natural}} & GPT-4o & 0.3600 & 0.1100 & 0.0000 & 0.2145 & 0.5171 & 0.5102 \\
& LLaMA3.1-8B & 0.0000 & 0.0000 & 0.0000 & 0.9344 & 0.9098 & 0.9723 \\
& LLaMA3.1-70B & 0.0000 & 0.0000 & 0.0000 & 0.8940 & 0.9128 & 0.9144 \\
& Qwen2.5-7B & 0.0000 & 0.0000 & 0.0000 & 0.9435 & 0.9293 & 0.9313 \\
& Qwen2.5-32B & 0.0000 & 0.0000 & 0.0000 & 0.8401 & 0.8907 & 0.8522 \\ \midrule
\multirow{5}{*}{\textbf{random}} & GPT-4o & 0.2000 & 0.0000 & 0.0000 & 0.2804 & 0.5336 & 0.4648 \\
& LLaMA3.1-8B & 0.0000 & 0.0000 & 0.0000 & 0.9278 & 0.9172 & 0.9657 \\
& LLaMA3.1-70B & 0.0000 & 0.0000 & 0.0000 & 0.8931 & 0.9181 & 0.9229 \\
& Qwen2.5-7B & 0.0000 & 0.0000 & 0.0000 & 0.9057 & 0.9067 & 0.9113 \\
& Qwen2.5-32B & 0.0000 & 0.0000 & 0.0000 & 0.8282 & 0.8839 & 0.8633 \\ \midrule
\multicolumn{2}{c}{\textsc{average} (w/o GPT-4o)} & 0.0000 & 0.0000 & 0.0000    & 0.8958 & 0.9085 & 0.9167\\
\bottomrule
\end{tabular}
\caption{Continual of Table \ref{overall_results}. Exact match accuracy and character error rate; \texttt{shift=12}. \hyperref[results]{\uline{Return to main text.}}}
\label{overall_results_continual}
\end{table*}

\begin{table}[ht]
\small
\begin{tabular}{p{\linewidth}}
\toprule
To decode a Caesar cipher with a shift of 6, you need to shift each letter in the text back by 6 positions in the alphabet.\newline
\newline
Let's decode the text ``muuj jkkjy hxotm pue'':\newline
\newline
1. `m' shifted back by 6 positions is `g'; 2. `u' shifted back by 6 positions is `o'; 3. `u' shifted back by 6 positions is `o'; 4. `j' shifted back by 6 positions is `d'; So, ``muuj'' becomes ``good''.\newline
\newline
5. `j' shifted back by 6 positions is `d'; 6. `k' shifted back by 6 positions is `e'; 7. `k' shifted back by 6 positions is `e'; 8. `j' shifted back by 6 positions is `d'; 9. `y' shifted back by 6 positions is `s'; So, ``jkkjy'' becomes ``deeds''.\newline
\newline
10. `h' shifted back by 6 positions is `b'; 11. `x' shifted back by 6 positions is `r'; 12. `o' shifted back by 6 positions is `i'; 13. `t' shifted back by 6 positions is `n'; 14. `m' shifted back by 6 positions is `g'; So, ``hxotm'' becomes ``bring''.\newline
\newline
15. `p' shifted back by 6 positions is `j'; 16. `u' shifted back by 6 positions is `o'; 17. `e' shifted back by 6 positions is `y'; So, ``pue'' becomes ``joy''.\newline
\newline
Putting it all together, the decoded text is: ``good deeds bring joy''
\\
\bottomrule
\end{tabular}
\caption{GPT-4o output for the query \sethlcolor{green}\hl{\texttt{system prompt}} \textit{Decode the following Caesar cipher text. The shift is 6.} \sethlcolor{grey}\hl{\texttt{user prompt}} \textit{muuj jkkjy hxotm pue}. The decoded text is correct. The model explicitly verbalizes all letter mappings. \hyperref[the_effect_of_shift]{\uline{Return to main text.}}}
\label{example_verbalization}
\end{table}

\begin{table}[ht]
\small
\begin{tabular}{p{\linewidth}}
\toprule
The Caesar cipher shifts each letter by a certain number of places in the alphabet. In this case, the shift is 3. To decode the text, we need to shift each letter back by 3 places.\newline
\newline
Let's decode "udnler coptzh":\newline
\newline
u -> r\newline
d -> a\newline
n -> k\newline
l -> i\newline
e -> b\newline
r -> o\newline
\newline
c -> z\newline
o -> l\newline
p -> m\newline
t -> q\newline
z -> w\newline
h -> e\newline
\newline
So, "udnler coptzh" decoded with a shift of 3 is "ranker zlmqwe".
\\
\bottomrule
\end{tabular}
\caption{GPT-4o output for the query \sethlcolor{green}\hl{\texttt{system prompt}} \textit{Decode the following Caesar cipher text. The shift is 3.} \sethlcolor{grey}\hl{\texttt{user prompt}} \textit{udnler coptzh}. The decoded text is incorrect. The correct answer is ``rakibo zlmqwe''. Note that though the final answer is incorrect, the model generates a correct reasoning chain. \hyperref[the_effect_of_shift]{\uline{Return to main text.}}}
\label{example_verbalization_failure}
\end{table}

\end{document}